\documentclass[10pt,twocolumn,letterpaper]{article}

\usepackage{iccv}
\usepackage{times}
\usepackage{epsfig}
\usepackage{graphicx}
\usepackage{amsmath}
\usepackage{amssymb}
\usepackage{booktabs}
\usepackage{multirow}
\usepackage{tikz}
\usepackage{pgfplots}
\usepackage{xcolor}
\usepackage[percent]{overpic}
\usepackage{caption}
\usepackage[accsupp]{axessibility}

\usepackage[breaklinks=true,bookmarks=false]{hyperref}

\iccvfinalcopy 

\def\iccvPaperID{6101} 

\ificcvfinal\fi

\begin{document}

\title{CROSSFIRE: Camera Relocalization On Self-Supervised Features from an Implicit Representation}


\author{
  Arthur Moreau$^{1,2}$, Nathan Piasco$^{1}$, Moussab Bennehar$^{1}$, Dzmitry Tsishkou$^{1}$\\ Bogdan Stanciulescu$^{2}$, Arnaud de La Fortelle$^{2}$\\
  $^{1}$Noah’s Ark IoV team, Huawei France,
  $^{2}$Mines Paris, PSL University, Centre for robotics\\
  \texttt{arthur.moreau2@huawei.com} \\}
\maketitle
\ificcvfinal\thispagestyle{empty}\fi
\begin{abstract}
   Beyond novel view synthesis, Neural Radiance Fields (NeRF) are useful for applications that interact with the real world. In this paper, we use them as an implicit map of a given scene and propose a camera relocalization algorithm tailored for this representation. The proposed method enables to compute in real-time the precise position of a device using a single RGB camera, during its navigation. In contrast with previous work, we do not rely on pose regression or photometric alignment but rather use dense local features obtained through volumetric rendering which are specialized on the scene with a self-supervised objective. As a result, our algorithm is more accurate than competitors, able to operate in dynamic outdoor environments with changing lightning conditions and can be readily integrated in any volumetric neural renderer.
\end{abstract}

\section{Introduction} \label{sec_introduction}

 Visual localization, i.e. the problem of camera pose estimation in a known environment~\cite{piasco2018survey}, enables to build camera-based positioning systems for various applications such as autonomous driving~\cite{coordinet}, robotics~\cite{sota_loc} or augmented reality~\cite{loc_ar}. Map-based navigation systems for such applications operate with a reference map of the environment, built from previously collected data. These maps are commonly defined with explicit 3D scenes representations (point cloud, voxels, meshes, etc.), which only store discrete information while the underlying environment they represent is continuous.

Recently, Neural Radiance Fields (NeRF)~\cite{nerf} and related volumetric-based approaches~\cite{instant-ngp, nerf_review} have emerged as a new way to implicitly represent a scene. 3D coordinates are mapped to volume density and radiance in a neural network. NeRF is trained with a sparse set of posed images of a scene and learns its 3D geometry via differentiable rendering. The resulting model is continuous, i.e. the radiance of all 3D points in the scene can be computed, which enables the rendering of photorealistic views from any viewpoint.

Beyond their rendering ability, implicit scene representations are actively investigated to be used as the map representation for navigation systems~\cite{nerf-nav,pantic2022sampling,lin2022mira, kwon2023renderable}. This work focuses on one aspect of the navigation pipeline, understudied in the specific case of implicit scene representation, the image localization problem. Our motivation is to provide a camera relocalization algorithm (i.e. 6-DoF pose estimation) from one RGB image based only on a learned volumetric-based implicit map. We aim to design a method for robotics applications: it must be fast to compute, robust to outdoor conditions and could be deployed in dynamic environments. Existing localization methods that use implicit maps either have limited accuracy by lack of geometric reasoning~\cite{lens,chen2022dfnet}, or do not meet the aforementioned requirements because photometric alignment~\cite{inerf,lin2022parallel} can be slow and assumes constant lightning conditions.

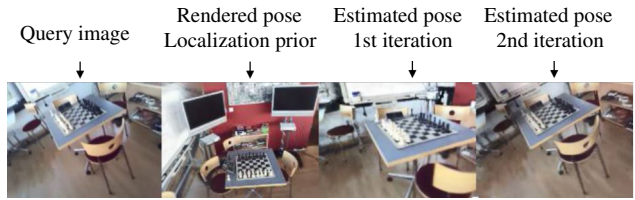
\begin{figure}[t]
   \centering
   \begin{overpic}[width=\linewidth]{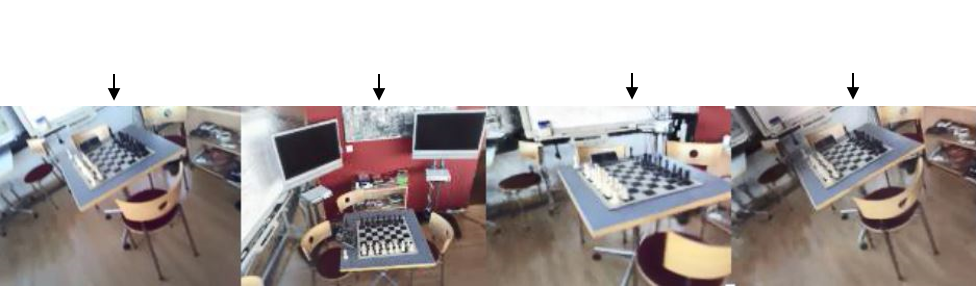}
   \put(2, 25){\footnotesize Query image}
   \put(27, 28){\footnotesize Rendered pose}
   \put(25, 24){\footnotesize Localization prior}
   \put(52, 28){\footnotesize Estimated pose}
   \put(55, 24){\footnotesize 1st iteration}
   \put(76, 28){\footnotesize Estimated pose}
   \put(78, 24){\footnotesize 2nd iteration}
   \end{overpic}
   \caption{\textbf{Visual localization in a neural renderer.} Starting from a coarse localization prior, our algorithm estimates the pose of a query image by comparing image features to descriptors rendered from a neural scene representation.}
   \label{fig:chess_prior}
\end{figure}

\paragraph{Contribution.} In this paper, we introduce local descriptors in NeRF's implicit formulation and we use the resulting model, named CROSSFIRE, as the scene representation of a 2D-3D features matching method. We train simultaneously a CNN feature extractor and a neural renderer to provide consistent scene-specific descriptors in a self-supervised way. During training, we leverage the 3D information learned by the radiance field in a metric learning optimization objective which does not require supervised pixel correspondences on image pairs nor a pre-computed 3D model. The proposed descriptors represent not only the local 2D image content but also the 3D position of the observed point, which enables to solve ambiguities in areas with repetitive patterns. Our method can use any differentiable neural renderer and, hence, can directly benefit from recent NeRF improvements. For instance, we make the model computationally tractable thanks to the multi-resolution hash encoding from Instant-NGP~\cite{instant-ngp} and adapted to dynamic outdoor scenes thanks to appearance embeddings from Nerf-W~\cite{nerfw}.

Finally, we show that these features can be used to solve the visual relocalization task with an iterative algorithm composed of a dense features matching step followed by standard Perspective-n-Points (PnP) camera pose computation. We take inspiration from structure-based visual localization pipelines~\cite{activesearch,sarlin2019coarse} but replace the commonly used sparse 3D model obtained from Structure-from-Motion by our neural field from which dense features are extracted. For a given camera pose candidate, we render dense descriptors and depth maps. Descriptors are used to establish 2D-2D matches which are upgraded to 2D-3D matches by the rendered depth. We can iteratively refine the estimated pose by repeating the aforementioned procedure, as presented in Figure~\ref{fig:chess_prior}.

\section{Related work} \label{sec_related}

\paragraph{Localization with Neural Scenes Representations.} Many algorithms have recently been developed to compute the camera pose of an image w.r.t. a NeRF model. 

One line of work has developed visual SLAM systems, where the implicit map is learned during the navigation. iMAP~\cite{imap} and NICE-SLAM~\cite{nice-slam} leverages the depth information of RGB-D cameras to de-couple pose and scene geometry estimation. Then, NeRF-SLAM~\cite{rosinol2022nerf} extends these approaches to RGB images by using dense monocular SLAM as supervision for the NeRF map. In contrast with these methods, we target a relocalization approach, where the environment has already been visited. In this scenario, the map is pre-computed offline or derived from a SLAM approach. Our solution could be used as a relocalization module that can be plugged into implicit SLAM pipelines for continuous navigation and place re-visit. 

A first relocalization solution is to align iteratively a query and a rendered image by optimizing the camera pose based on the photometric error. This has been first proposed by iNeRF~\cite{inerf} which demonstrates accurate pose estimation on usual NeRF datasets, i.e. controlled environments such as synthetic or static indoor scenes. However, the localization process is slow because each iteration requires rendering and backpropagation through the entire NeRF model, and the convergence basin is small. This idea has then been improved by using more efficient rendering models and parallel optimization based on Monte-Carlo sampling~\cite{lin2022parallel}. Loc-NeRF~\cite{maggio2023loc} integrates this idea in a particle filter formulation.

Another direction uses Absolute Pose Regression~\cite{PoseNet,coordinet} that directly connects images and camera poses in a deep network. While these methods usually present a low accuracy~\cite{Sattler2019}, they can be improved by leveraging a NeRF during the training step. Direct-PoseNet~\cite{chen2021direct} renders the image at the estimated pose and uses the differentiability of the renderer to define an additional loss function based on the photometric error. Then, DF-Net~\cite{chen2022dfnet} iterates on this idea and defines a loss based on features matching. Finally, LENS~\cite{lens} pre-computes a large set of synthetic views uniformly distributed across the scene and uses it as additional training data. 


Related to our work, Features Query Network~\cite{FQN} stores local descriptors in an implicit scene representation and uses it to perform local features matching in a structure-based formulation~\cite{activesearch,sarlin2019coarse,Panek2022ECCV}. While we use a related localization process, our method is novel on two crucial aspects. First, FQN is limited to a pre-computed sparse 3D point cloud, while our proposal provides dense features from a radiance field. Then, instead of memorizing in a supervised way how descriptors vary w.r.t. viewpoint in an off-the-shelf features extractor, we take the opposite direction and learn scene-specialized descriptors without supervision through a metric learning objective and decide to model these features as not dependent on the viewing direction, in order to facilitate the matching process. To the best of our knowledge, learning visual localization descriptors in a neural radiance field without supervision has not been proposed before.

\paragraph{Learning-based description of local features.} Local descriptors provide useful descriptions of regions of interest that enable to establish accurate correspondences between pairs of images describing the same scene. While hand-crafted descriptors such as SIFT~\cite{lowe1999object, lowe2004distinctive} and SURF~\cite{bay2006surf} have known great success, the focus has shifted in recent years to learn features extraction from large amounts of visual data. Many learning-based formulations~\cite{UCN2016, jahrer2008learned, simo2015discriminative, zagoruyko2015learning, tian2019sosnet, Dusmanu2019CVPR} rely on siamese convolutional networks trained with pairs or triplets of images/patches supervised with correspondences. NeRF-Supervision~\cite{yen2022nerfsupervision} takes advantage of the geometric consistency of depth-supervised object-centric NeRFs to obtain correspondences between different views of the object in order to learn view-invariant dense object descriptors. Features extractors can be trained without annotated correspondences by augmenting two versions of a same image or using weak supervision. SuperPoint~\cite{detone18superpoint} uses homographies while Novotny et al.~\cite{novotny2018self} leverage image warps. In a recent work, CAPS~\cite{wang2020learning} have shown that accurate correspondences between different views can be obtained using weak supervision through the use of relative camera poses. Our proposed method follow a different path to learn repeatable descriptors: we constraint the image feature extractor to provide the same descriptors map as the Neural Field. This approach allows us to learn dense scene-specific descriptors without annotated correspondences since the neural renderer provides similar features for rays which intersect the same point.



\section{Method} \label{sec_method}

The proposed algorithm estimates the 6-DoF camera pose of a query image in an already visited environment. We first train our modules in an offline step, using a set of reference images with corresponding poses, captured beforehand in the area of interest. A 3D model of the scene is not a pre-requisite because we learn the scene geometry during the training process.

\begin{figure}[t]
   \centering
   \begin{overpic}[width=.8\linewidth]{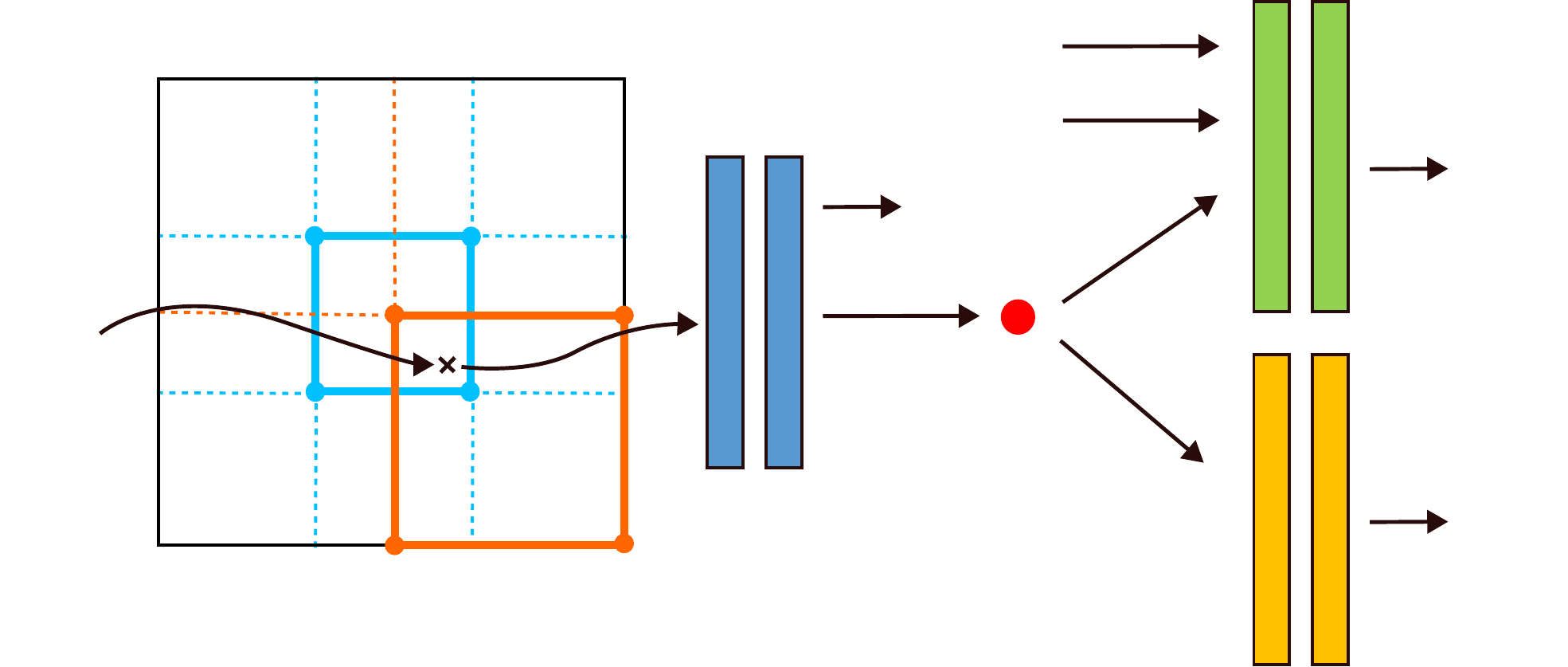}
   \put(-5, 18){\small XYZ}
   \put(58, 28.5){\small $\sigma$}
   \put(64, 38){\small d}
   \put(59, 33){\footnotesize $\mathcal{L}_{i}^{(a)}$}
   \put(93, 30){\small RGB}
   \put(93, 7.5){\small Descriptor}
   \end{overpic}
   \caption{\textbf{Neural radiance and descriptors fields.} The input coordinate is encoded by the multi-resolution hash tables from Instant-NGP~\cite{instant-ngp} enabling fast training and rendering. We use per-image appearance embeddings to handle varying illumination across training images. The descriptors heads is invariant to viewing direction and appearance vector allowing to learn robust localization features.}
   \label{neural renderer}
\end{figure}
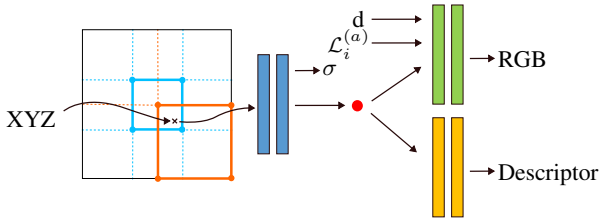

\subsection{Neural rendering of descriptors}

\paragraph{Background.} NeRF~\cite{nerf} is capable of rendering a view from any camera pose in a given scene while being trained only with a sparse set of observations. Given a camera pose with known intrinsics, 2D pixels are back-projected in the 3D scene through ray marching. The density $\sigma$ and RGB color $c$ of each point $p=(x,y,z)$ along the ray are evaluated by a MLP $R_{\theta}$: $c, \sigma = R_{\theta}(p,d)$ where $d$ is the viewing direction. The final pixel color of a pixel is computed with differentiable volumetric rendering along the ray, which enables to train the implicit scene representation by minimizing the photometric error of rendered images.

NeRF makes the assumption that illumination in the scene remain constant over time, which does not hold for many real world scenes. NeRF-W~\cite{nerfw} overcomes this limitation by modeling appearance with a per-image latent codes $\mathcal{L}_{i}^{(a)}$ (i.e. appearance embedding) that controls the appearance of each rendered view. Another limitation the original formulation of NeRF is the computation time: rendering an image requires $H\times W \times N$ evaluations of the 8 layers MLP, where $N$ is the number of points sampled per ray, resulting in slow training and rendering. Recently, Instant-NGP~\cite{instant-ngp} proposes to use multi-resolution hash encoding to accelerate the process by storing local features in hash tables, which are then processed by much smaller MLPs compared to NeRF resulting in significant improvement of both training and inference times.

\paragraph{Neural radiance and descriptors fields.} CROSSFIRE combines the 3 aforementioned techniques to efficiently render dynamic scenes. However, our main objective is not photorealistic rendering but, rather, features matching with new observations. While it is possible to align a query image with a NeRF model by minimizing the photometric error~\cite{inerf}, such approach lacks robustness w.r.t. variations in illumination. Instead, we propose to add positional features, i.e. $D$-dimensional latent vectors which describe the visual content of a region of interest in the scene, as an additional output of the radiance field function. In contrast with the rendered color, we model these descriptors as invariant to viewing direction $d$ and appearance vector $\mathcal{L}_{i}^{(a)}$ (\textit{i.e.} we do not provide $d$ and $\mathcal{L}_{i}^{(a)}$ to the MLP head responsible of generating the positional feature, see Figure~\ref{neural renderer}). We verify through ablation study in section~\ref{subsec_ablation} that this descriptor property makes the matching process more robust. Similar to color, the 2D descriptor of a camera ray is aggregated by the usual volumetric rendering formula applied on descriptors of each point along the ray. The architecture of our proposed neural renderer is summarized in Figure~\ref{neural renderer} and implementations details are provided in section. The training pipeline of CROSSFIRE is explained in the next section.

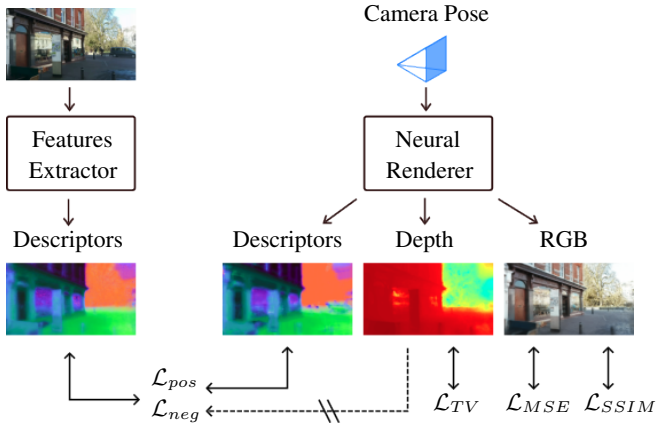
\begin{figure}[t]
   \centering
   \begin{overpic}[width=\linewidth]{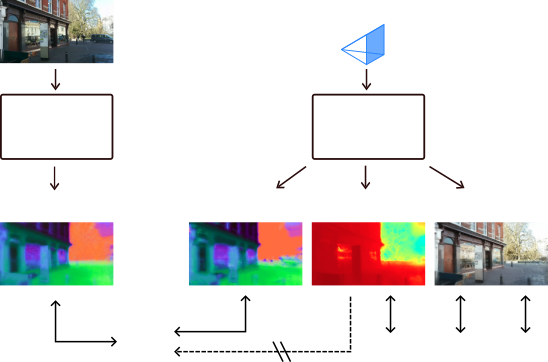}

    \put(57, 64){\small Camera Pose}

    \put(1, 28){\small Descriptors}
    \put(36, 28){\small Descriptors}
    \put(62, 28){\small Depth}
    \put(85, 28){\small RGB}

    \put(23, 6){\small $\mathcal{L}_{pos}$}
    \put(23, 1){\small $\mathcal{L}_{neg}$}
    \put(68, 2){\small $\mathcal{L}_{TV}$}
    \put(79.5, 2){\small $\mathcal{L}_{MSE}$}
    \put(92, 2){\small $\mathcal{L}_{SSIM}$}

    \put(4, 44){\small Features}
    \put(3.5, 39){\small Extractor}

    \put(62, 44){\small Neural}
    \put(60.5, 39){\small Renderer}
    
   \end{overpic}
   \caption{\textbf{Training pipeline of CROSSFIRE.} We jointly optimize the neural renderer and the features extractor to obtain robust, scene-specific localization descriptors. We use regularization losses (i.e. TV and SSIM) to increase the consistency of the neural renderer. We propose a two-terms loss that maximizes the similarity between corresponding feature maps while penalizing pixel pairs that are geometrically distant from each other.}
   \label{training}
\end{figure}

\subsection{Self-supervised training of features}

\paragraph{Motivation.} In the previous section, we explained how our proposed neural renderer describes the map for relocalization purposes thanks to the introduced positional descriptors. Additionally, we also need to extract features from the query image. A simple solution, proposed by FQN~\cite{FQN}, is to use an off-the-shelf pre-trained features extractor such as SuperPoint~\cite{detone18superpoint} or D2-Net~\cite{Dusmanu2019CVPR}, and train the neural renderer to memorize observed descriptors depending on the viewing direction. Optimizing scene-specific descriptors, however, allows to better differentiate repetitive patterns in the scene resulting, in a more robust localization and reducing failure cases. To this end, we propose to train jointly the feature extractor with the neural renderer by defining an optimization objective which leverages the scene geometry. We obtain descriptors specialized on the target scene which describe not only the visual content but also the 3D location of the observed point, with better discriminant property than generic descriptors.

\vspace{5pt}
The training procedure of our system is described in Figure~\ref{training}. One training sample corresponds to a reference image with its corresponding camera pose. From one side, the image is processed by the features extractor to obtain the descriptors map $F_{I}$. On the other side, we sample points along rays for each pixel using camera intrinsics, compute density, color and descriptor of each 3D point, and finally perform volumetric rendering to obtain a RGB view $C_{R}$, a descriptors map $F_{R}$ and a depth map $D_{R}$.

\paragraph{Features Extraction.} Our features extractor, inspired by SuperPoint~\cite{detone18superpoint}, is a simple fully convolutional neural network with 8 layers, ReLU activations and max poolings. The input is a RGB image $I$ of size $H \times W$ and produces a dense descriptors map $F_{I} \in \mathbb{R}^{H/4 \times W/4 \times d}$.

\paragraph{Learning the Radiance Field.} Similar to NeRF~\cite{nerf}, we use the mean squared error loss  $\mathcal{L}_{MSE}$ between $C_{R}$ and the real image to learn the radiance field. As we render entire, although downscaled, images in a single training step, we can leverage the local 2D image structure and minimise the structural dissimilarity (DSSIM) loss $\mathcal{L}_{SSIM}$~\cite{ssim}, which we observe to produce sharper images and more accurate scene geometry. Depth maps are used by the localization process to compute the camera pose, and then better depth results in more accurate poses. NeRF models trained with limited training views can yield incorrect depths, due to the shape-radiance ambiguity~\cite{kaizhang2020}. We add a regularization loss $\mathcal{L}_{TV}$ which minimizes depth total variation of randomly sampled 5x5 image patches to encourage smoothness and limit artefacts on the rendered depth maps~\cite{niemeyer2022regnerf}. We verify in section~\ref{subsec_ablation} that using these 3 loss functions is beneficial for the localization accuracy.

\begin{figure}[t]
   \centering
   \includegraphics[width=\linewidth]{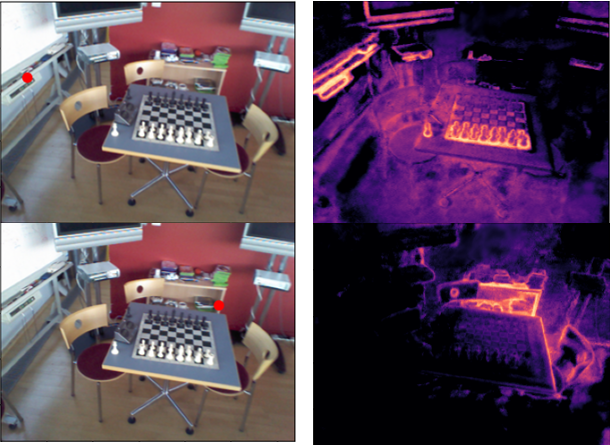}
   \caption{\textbf{Similarities of positional features.} We show the dense matching map between one descriptor from the query image (red dots in left images) and the reference descriptors from the neural renderer. Thanks to our training objective, descriptors close (in 3D) to the selected points have high similary whereas others do not match. This behaviour is enforced by our loss function.}
   \label{dense_matching}
\end{figure}

\paragraph{Learning the Descriptors Field.} Our main goal is to match the descriptors map from the CNN features extractor and the corresponding one from the neural renderer. The self-supervised optimization objective encourages both models to produce identical features for a given pixel while preventing high matching scores between points far from each other in the 3D scene. We define a loss function with two terms $\mathcal{L}_{pos}$ and $\mathcal{L}_{neg}$, applied on a pair of descriptors maps, each containing $n$ pixels. We use the cosine similarity, noted $\otimes$, to measure similarity between descriptors.

The first loss term $\mathcal{L}_{pos}$ maximizes the similarity between descriptors maps $F_{I}$ and $F_{R}$ from both models:

\begin{equation}
\mathcal{L}_{pos} = \frac{1}{n} \sum_{i=1}^{n} \text{max}(0, 1 - F_{I}[i] \otimes F_{R}[i])
\end{equation}

The second loss term $\mathcal{L}_{neg}$ samples random pairs of pixels and ensures that pixel pairs with large 3D distances have dissimilar descriptors:

\begin{equation}
\mathcal{L}_{neg} = \frac{1}{mn} \sum_{k,i=1}^{m,n} \text{max}(0, F_{I}[p_{k}(i)] \otimes F_{R}[i] - t_{\lambda}(p_{k}(i),i))
\end{equation}
where $t_{\lambda}(i,j) = max(0, 1 - \lambda \| xyz(i) - xyz(j) \| )$. $xyz(i)$ is the 3D coordinate of the point represented by the $i_{th}$ pixel in the descriptors map. We compute it from the camera parameters of the rendered view and predicted depth. It should be noted that we do not backpropagate the gradient of this loss to the depth map because the gradient of this loss does not provide meaningful signal to learn the scene geometry. $\lambda$ is an hyperparameter which controls the maximum similarity between descriptors at a given 3D distance. $(p_{k})_{m}$ are random permutations of pixel indices from 1 to n.

The proposed self-supervised objective is close to a classical triplet loss~\cite{netvlad}, but we show in Figure ~\ref{abl_triplet} that scaling the loss by the 3D coordinates in the formulation is crucial to learn smooth and selective descriptors. A visualization of the similarity between descriptors enforced by the proposed loss is shown in Figure ~\ref{dense_matching}.

Finally, we optimize the following loss function at each training step:

\begin{equation}
    \mathcal{L} = \mathcal{L}_{MSE} + \lambda_1\mathcal{L}_{SSIM} + \lambda_2\mathcal{L}_{TV} + \mathcal{L}_{pos} + \mathcal{L}_{neg}
\end{equation}
where $\lambda_1 = 0.1$ and $\lambda_2 = 1e^{-3}$ are hyper-parameters introduced to balance SSIM and TV losses, respectively.


\vspace{15pt}

\subsection{Visual Localization by iterative dense features matching}

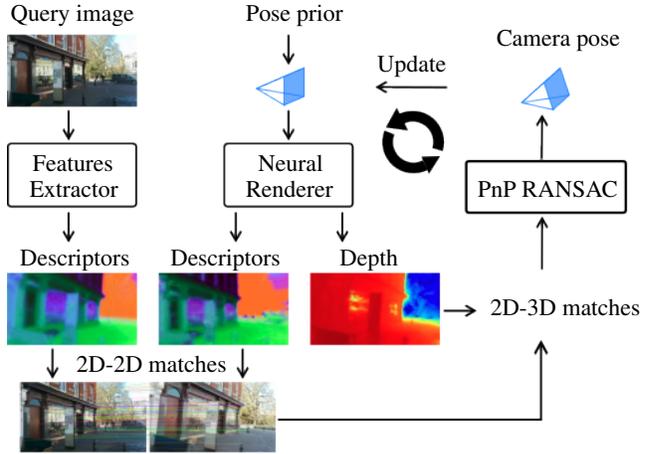
\begin{figure}[t]
   \centering
   \begin{overpic}[width=\linewidth]{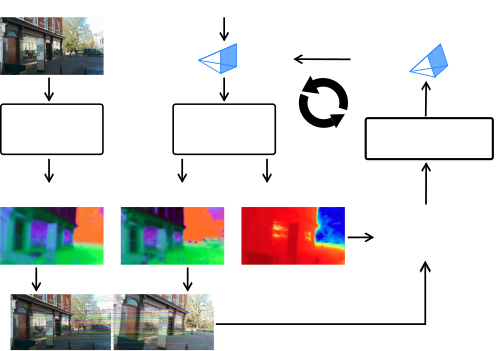}
   \put(0.5, 69){\small Query image}
   \put(38, 69){\small Pose prior}
   \put(78, 65){\small Camera pose}

   \put(59, 61){\small Update}

   \put(4, 45){\small Features}
   \put(3.5, 41){\small Extractor}
   \put(40, 45){\small Neural}
   \put(38, 41){\small Renderer}
   \put(75, 41){\small PnP RANSAC}

   \put(2, 30){\small Descriptors}
   \put(26, 30){\small Descriptors}
   \put(53, 30){\small Depth}

   \put(77, 22){\small 2D-3D matches}

   \put(11, 13){\small 2D-2D matches}
   
   \end{overpic}
   \caption{\textbf{CROSSFIRE localization procedure.} Descriptors are extracted from the query image and matched against descriptors rendered from the localization prior. Depth information provides 2D-3D matches that enable to compute the pose with PnP + RANSAC. This process can be repeated iteratively, by rendering descriptors from the predicted pose.}
   \label{localization_figure}
\end{figure}

This section describes the localization pipeline used to estimate the camera pose of a given query image using our learned renderer and features. An overview of this procedure is shown in Figure~\ref{localization_figure}. The proposed solution combines simple and commonly used techniques and we do not claim algorithmic novelty on this part. The goal is, rather, to demonstrate that the quality and robustness of our learned features enables to reach precise localization while using basic features matching and pose estimation strategies.

\paragraph{1. Localization prior.} Similar to related features matching methods~\cite{activesearch, sarlin2019coarse, FQN}, we assume to have access to a localization prior, i.e. a camera pose relatively close to the query pose. A view observed from the prior should have an overlapping visual content with the query image to make the matching process feasible. Such priors can be obtained by matching a global image descriptor against an image retrieval database~\cite{netvlad, sarlin2019coarse} or an implicit map~\cite{ImPosing}.

\paragraph{2. Features extraction.} First, we extract dense descriptors from the query image through the CNN. On the other side, descriptors and depth corresponding to the localization prior are computed by the neural renderer.

\paragraph{3. Dense Features Matching.} Query and reference descriptors are matched with cosine similarity. We consider that 2 descriptors are a match if the similarity is higher than a threshold $\theta$ and if it represent the best candidate in the other map in both direction (mutual matching). We then compute the predicted 3D coordinate of rendered pixels which have been matched (thanks to camera parameters and depth) and obtain a set of 2D-3D matches.

\paragraph{4. Camera Pose Estimation.} To compute the camera pose from the 2D-3D matches, we use the Perspective-N-Points algorithm combined with RANSAC~\cite{RANSAC}, in order to get a robust estimate by discarding outliers matches.

\paragraph{5. Iterative Pose Refinement.} While classical 3D models only have access to a finite set of reference descriptors, our neural renderer can compute them from any camera pose. Similar to FQN~\cite{FQN} and ImPosing~\cite{ImPosing}, we can then consider the camera pose estimate as a new localization prior and iterate the previously mentioned steps multiple times to refine the camera pose.

\section{Experiments} \label{sec_experiments}

\begin{figure}[t]
   \centering
   \includegraphics[width=\linewidth]{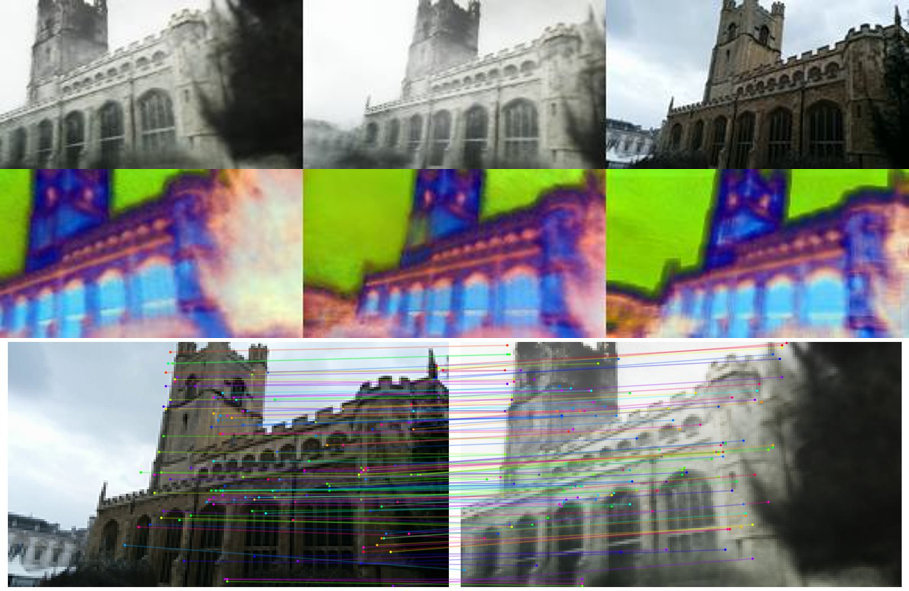}
   \caption{\textbf{Visualization of rendered views, descriptors and matches in StMarysChurch.} We show on the top row the query image (right), the RGB rendered view from the localization prior (left) and from the 1st estimated pose (middle). The second row represents a PCA visualization of the corresponding descriptors map from the neural renderer (left and middle) and the features extractor (right). The last row displays the inlier matches obtained by our pipeline.}
   \label{big_fig}
\end{figure}

\begin{figure*}[h]
    \centering
    \includegraphics[width=\textwidth]{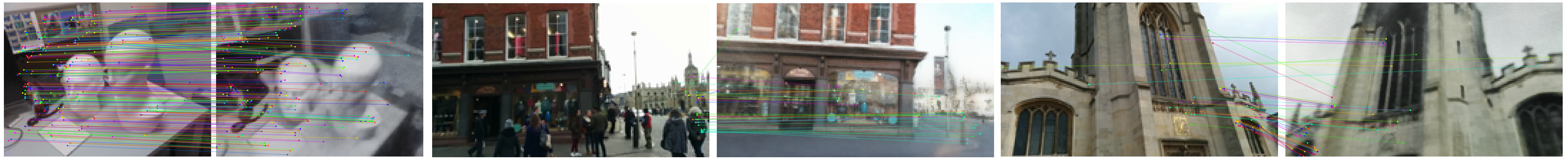}
    \caption{\textbf{Success and failure cases:} we show inliers matches between the query image and the NeRF rendered image at prior pose. Using dense features field for localization enables to establish accurate correspondences in texture-less areas (left). Failure cases are observed in the presence of dynamic objects (middle), for which the PnP converges on a wrong pool of matches, and ambiguous cases (right) where the CNN mixes up the symmetrical parts of the church due to lack of long-range reasoning.}
    \label{fig:npff_quali}
\end{figure*}

We first present a comparison of CROSSFIRE with related methods relocalization that use implicit map representations in section~\ref{sec:sota}. We also evaluate the impact of the localization prior in section~\ref{sec:npff_prior} and additional ablation studies in~\ref{subsec_ablation}. 

\paragraph{Implementation.} Our system is implemented in PyTorch. The hash tables and MLPs of the neural renderer use tiny-cuda-nn~\cite{tiny-cuda-nn}. We use the default PnP pose solver from PoseLib~\cite{PoseLib}. In all the proposed experiments, we use descriptors of size 32. We train the models for 100k iterations. The initial learning rate is set to $1e^{-3}$ and reduced to $1e^{-4}$ after 2000 iterations. For ensuring reproducibility, the detailed architecture of our neural networks are provided in supplementary materials.

\paragraph{Datasets.} We evaluate our method on 2 standard localization benchmarks. 7scenes~\cite{7scenes} consists in indoor static scenes captured using a hand-held camera. Cambridge Landmarks~\cite{PoseNet} contains outdoor scenes representing buildings observed from different viewpoints and lighting conditions, with dynamic occluders such as pedestrians and cyclists in both train and test sets.  

\paragraph{Efficiency.} The storage requirement of our modules is 50MB (48MB for the hash tables and 2MB for the neural networks). In contrast with explicit maps, this number does not grow with the amount of reference data. All trainings and inferences have been performed on a RTX3090 GPU. Trainings take approximately 5 hours for indoor scenes and 15 hours for larger outdoor scenes. Inference times are: 9ms for features extraction, 5ms for rendering, 5ms for dense matching and $\approx60$ms for PnP+RANSAC (because we have a lot of matches), resulting in $\approx200$ms for the total time with 3 iterations reported in the experiments. Speedup can be achieved easily by less refinements, at the cost of minor accuracy drop.

\subsection{Comparison to related methods}
\label{sec:sota}
We evaluate our method on both datasets using a maximum of 3 iterations of the localization process. We use as localization prior the top 1 reference pose retrieved by DenseVLAD~\cite{densevlad}. In order to render reference frames efficiently, the matching step is done at a small resolution: 194x108 for Cambridge Landmarks and 161x120 for 7scenes.

\begin{table*}[t]
\centering
\small
\begin{tabular}{|l|ccc|ccc|}
\hline
Dataset / Methods & \multicolumn{3}{c|}{\hfil Absolute Pose Regression + NeRF} & \multicolumn{3}{c|}{\hfil Implicit local features}\\ \hline
Cambridge & \multicolumn{1}{p{1.5cm}}{DirectPN~\cite{chen2021direct}} & \multicolumn{1}{p{1.5cm}}{DFNet \cite{chen2022dfnet}} & \multicolumn{1}{p{1cm}|}{LENS~\cite{lens}} &  \multicolumn{1}{p{2cm}}{FQN-D2N~\cite{FQN}} & \multicolumn{1}{p{1.9cm}}{FQN-MN~\cite{FQN}} & \multicolumn{1}{p{2.8cm}|}{\textbf{CROSSFIRE (Ours)}} \\ \hline
Kings College & - & 0.73m / 2.4° & 0.33m / 0.5° & 0.32m / 0.5° &  \textbf{0.28m / 0.4°}  & 0.47m / 0.7°  \\
Old Hospital & - & 2.00m / 3.0° & 0.44m / 0.9° & 0.64m / 0.9° & 0.54m / 0.8° &  \textbf{0.43m / 0.7°} \\
Shop Facade  & - & 0.67m / 2.2° & 0.27m / 1.6° & 0.14m / 0.6° &  \textbf{0.13m / 0.6°} & 0.20m / 1.2° \\
StMarys Church & - & 1.37m / 4.0° & 0.53m / 1.6° & 0.93m / 3.5° & 0.58m / 2.0° &  \textbf{0.39m / 1.4}° \\ \hline
Average & - & 1.19m / 2.9° & 0.39m / 1.2° & 0.51m / 1.4° & 0.38m / 1.0° &  \textbf{0.37m / 1.0°} \\ \hline
7scenes & & & & & & \\ \hline
Chess  & 0.10m / 3.5° & 0.05m / 1.9° & 0.03m / 1.3° & 0.06m / 1.9° & 0.04m / 1.3° & \textbf{0.01m / 0.4°} \\
Fire & 0.27m / 11.7° & 0.17m / 6.5° & 0.10m / 3.7° & 0.14m / 4.1° & 0.10m / 3.0° & \textbf{0.05m / 1.9°} \\
Heads & 0.17m / 13.1° & 0.06m / 3.6° & 0.07m / 5.8° & 0.05m / 3.5° & 0.04m / 2.4° & \textbf{0.03m / 2.3°} \\
Office & 0.16m / 6.0° & 0.08m / 2.5° & 0.07m / 1.9° & 0.14m / 4.1° & 0.10m / 3.0° & \textbf{0.05m / 1.6°} \\
Pumpkin & 0.19m / 3.9° & 0.10m / 2.8° & 0.08m / 2.2° & 0.10m / 2.6° & 0.09m / 2.4° & \textbf{0.03m / 0.8°} \\
Kitchen & 0.22m / 5.1° & 0.22m / 5.5° & 0.09m / 2.2° & 0.18m / 4.8° & 0.16m / 4.4° & \textbf{0.02m / 0.8°} \\
Stairs & 0.32m / 10.6° & 0.16m / 3.3° & 0.14m / 3.6° & 1.41m / 53.0° & 1.40m / 34.7° & \textbf{0.12m / 1.9°} \\ \hline
Average & 0.20m / 7.3° & 0.12m / 3.7° & 0.08m / 3.0° & 0.30m / 10.6° & 0.28m / 7.3° & \textbf{0.04m / 1.1°} \\ \hline
\end{tabular}
\caption{\label{tab:sota_results} \textbf{6-DoF median localization errors of visual localization methods based on implicit representations.} Direct-PoseNet did not report results for Cambridge Landmarks.}
\end{table*}

We compare our algorithm to the learning-based visual relocalization methods that use implicit map representations in their pipeline.
\begin{itemize}
    \setlength\itemsep{1em}
    \item Direct-PoseNet~\cite{chen2021direct} train an Absolute Pose Regressor with an additional photometric loss by rendering the estimated pose through NeRF.
    \item DFNet~\cite{chen2022dfnet} goes in the same direction but defines a features matching loss with the rendered view.
    \item LENS~\cite{lens} trains an absolute pose regressor with NeRF rendered views uniformly distributed across the scene.
    \item FQN~\cite{FQN} regresses descriptors in an implicit representation of a sparse 3D model. This method is the closest to our work because it uses the same iterative localization process and store descriptors in a neural scene representation. The main differences are that descriptors are not trained specifically from the scene but memorized from a pretrained features extractors, and that the representation is sparse whereas ours is dense. Results are reported for D2-Net~\cite{Dusmanu2019CVPR} and MobileNetv2~\cite{sandler2018mobilenetv2} descriptors.
\end{itemize}

iNerf~\cite{inerf} and related methods are not present in our evaluation, first because results on usual localization benchmarks are not reported in the corresponding papers, but also because it does not meet the robotics requirements described before, i.e. fast inference for iNeRF and compatibility with outdoor dynamic environments.

The results of the comparisons for both datasets are shown in Table~\ref{tab:sota_results}. CROSSFIRE obtains the lowest error for both indoor localization and  outdoor scenes. Results on the highly ambiguous Stairs scene are higher than in other scenes but still better than other methods for which the localization process sometimes totally fail.

Furthermore, we consistently perform better than NeRF-assisted APR methods and, more importantly, than pretrained implicit descriptors. Because the camera pose estimation process used in FQN is similar than in ours, these results indicate that our scene-specific features are beneficial compared to off-the-self features extractors. 

We hypothesize that the absolute localization accuracy in outdoor scenes is lower for 2 main reasons. First, we lack a way to handle dynamic content such as pedestrians during the test step, which we observe to degrade the quality of our matches. Second, the quality of depth maps in these scenes is less accurate than in indoor scenarios, especially for background, due to observable image content very far from the camera. As we use depth to compute the 3D coordinates of matches, this introduces noise in the localization process.

\subsection{How good the pose priors need to be?}
\label{sec:npff_prior}

To measure how bad initialization impacts localization results, we conducted an experiment on the Chess scene where we replace the prior from image retrieval by using the same prior for all test images (shown in  Figure~\ref{fig:chess_prior}). Results are shown in Table~\ref{tab:prior}. We observe that, thanks to our iterative refinement, imprecise priors do not affect the final localization accuracy but rather require more iterations to reach the correct camera pose.

\begin{table}[h]
    \small
    \centering
    \begin{tabular}{|l|l|l|l|l|}
    \hline
        cm / ° & Prior & Iter 1 & Iter 2 & Iter 3  \\ \hline
        Retrieval & 0.22 / 12.1 & 0.02 / 0.7 & 0.01 / 0.5 & 0.01 / 0.4  \\ \hline
        Constant & 1.82 / 32.2 & 0.12 / 2.8 & 0.02 / 0.6 & 0.01 / 0.5  \\ \hline
    \end{tabular}
    \caption{\textbf{Impact of prior accuracy:} Median error w.r.t. prior strategy and iterations.}
    \label{tab:prior}
\end{table}

\subsection{Evaluation of the features extractor}

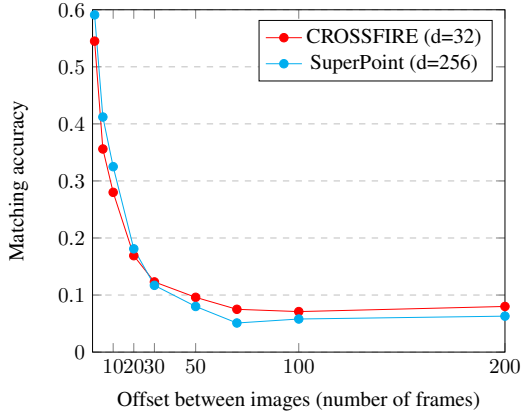
\begin{figure}[t]
\begin{tikzpicture}[xscale=0.8,yscale=0.8]
\begin{axis}[
    xlabel={Offset between images (number of frames)},
    ylabel={Matching accuracy},
    xmin=0, xmax=200,
    ymin=0, ymax=0.6,
    xtick={10,20,30,50,100,200},
    ytick={0,0.1,0.2,0.3,0.4,0.5,0.6},
    legend pos=north east,
    ymajorgrids=true,
    grid style=dashed,
]
\addplot[
    color=red,
    mark=*,
    ]
    coordinates {
    (1,0.545)(5,0.356)(10,0.28)(20,0.169)(30,0.123)(50,0.096)(70,0.075)(100,0.071)(200,0.080)
    };

\addplot[
    color=cyan,
    mark=*,
    ]
    coordinates {
    (1,0.591)(5,0.412)(10,0.325)(20,0.181)(30,0.117)(50,0.08)(70,0.051)(100,0.058)(200,0.063)
    };
    
\legend{CROSSFIRE (d=32), SuperPoint (d=256)}
\end{axis}
\end{tikzpicture}
\caption{\textbf{Comparison between features extractors.} We plot the matching accuracy on the Chess scene depending on the time offset between images.} 
\label{fig:features_extractor}
\end{figure}

Beyond the localization accuracy of the entire method, we conducted an experiment to compare the matching accuracy of our scene-specialized features extractor to those of SuperPoint~\cite{detone18superpoint}, which is a popular pre-trained learning-based method. Because we need to train a neural field on each scene, we can't use the HPatches benchmark~\cite{hpatches_2017_cvpr} for such purpose.
On the Chess scene, we compute reference matches between test images thanks to NeRF geometry. Then, we compare it with predicted matches to compute the matching accuracy (or precision). We compute this for image pairs with varying time offsets and report results in Figure~\ref{fig:features_extractor}. We observe that SuperPoint descriptors enable better matching when the viewpoint is close, but CROSSFIRE is more accurate with large viewpoint discrepancy. It should be noted that CROSSFIRE descriptors are 8 times more compact (32 vs 256).

\subsection{Qualitative evaluation}

We provide visualization of success and failure cases in Figure~\ref{fig:npff_quali}. In the first failure case in the Shop Facade scene (middle), we observe that the set of inliers matches is entirely incorrect, but, probably out of bad luck, consistently lead to a (wrong) camera pose. The RANSAC loop selected this pool of correspondences that lies in pedestrians instead of other matches on the shop. This problem could be addressed by confidence estimation, that we leave as future work.
For the second case, the only way to distinguish the left side from the right side of the church is to reason on the entire image, since symmetrical parts are locally similar. Because the confusing areas are far from each other in the image and our CNN uses small convolutional filters, such long-range reasoning is prevented and features from the right side in the query are wrongly matched with the left side. This could be improved with attention mechanisms in the features extractor architecture.

More visualizations are provided in Figure~\ref{big_fig} and in the supplementary video.

\subsection{Ablation studies}
\label{subsec_ablation}

\begin{figure}[t]
   \centering
   \includegraphics[width=\linewidth]{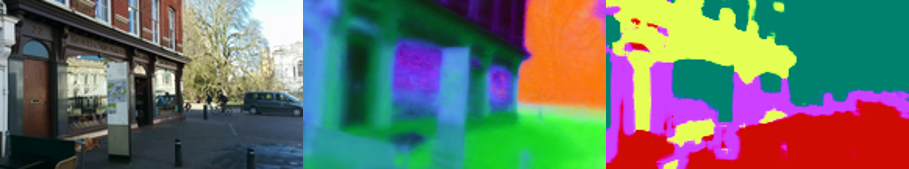}
   \caption{\textbf{Qualitative comparison of descriptors between the proposed loss and a classical triplet loss.} We visualize the PCA of descriptors from our loss (middle) and a triplet (right) for a given query image (left).}
   \label{abl_triplet}
\end{figure}

\paragraph{Descriptor loss.} The self-supervised loss used to train descriptors is similar to the triplet loss commonly used for metric learning, except an additional term for negative pairs which depends on the 3D distance between points. We propose a qualitative comparison between the triplet loss and our proposal in Figure~\ref{abl_triplet}. We observe that the representation learned by our system is smooth and more expressive than the triplet loss which only separate the scene into few clusters. More details including a quantitative comparison is provided in supplementary materials.

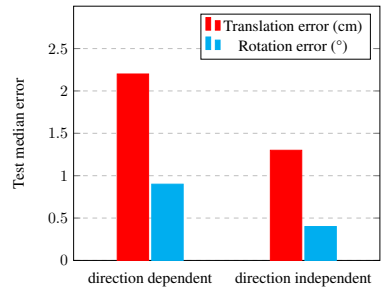
\begin{figure}[b]
    \centering
    \resizebox{0.6\columnwidth}{!}{%

        \begin{tikzpicture}
        \begin{axis}[major x tick style = transparent,
            ybar,
            ylabel={Test median error},
            ymin=0, ymax=3,
            enlarge x limits=0.5,
            bar width=20pt,
            ymajorgrids = true,
            symbolic x coords={direction dependent,direction independent},
            ytick={0,0.5,1.0,1.5,2.0, 2.5},
            legend pos=north east,
            grid style=dashed,
            xtick = data
        ]

        \addplot[color=red, fill=red]
            coordinates {(direction dependent,2.2)(direction independent,1.3)};
        
        \addplot[color=cyan, fill=cyan]
            coordinates {
            (direction dependent,0.9)(direction independent,0.4)
            };
            
        \legend{Translation error (cm),Rotation error (°)}
        \end{axis}
        \end{tikzpicture}
    }
    \caption{\textbf{Localization accuracy depending on descriptor head inputs.} We compare the final accuracy on the ``Chess" scene with and without the viewing direction as descriptor input in the neural renderer.}
    \label{fig:abl_direction}
\end{figure}

\paragraph{Conditioning descriptors with viewing direction.} We modeled the descriptors learned by the neural renderer as independent of the direction from which the point is observed. We verify that this choice is relevant by comparing it to the view-dependent case. Modeling the descriptors as dependent on the image appearance is not feasible because this parameter is unknown during the localization step. The comparison is shown in Figure~\ref{fig:abl_direction}.

\begin{figure}[t]
    \centering
    \resizebox{0.7\columnwidth}{!}{%

        \begin{tikzpicture}
        \begin{axis}[major x tick style = transparent,
            ybar,
            ylabel={Test median error},
            ymin=0, ymax=8,
            enlarge x limits=0.2,
            bar width=20pt,
            ymajorgrids = true,
            symbolic x coords={MSE+SSIM, MSE+TV,MSE+TV+SSIM},
            ytick={0,1.0,2.0, 3.0,4.0,5.0,6.0,7.0 },
            legend pos=north east,
            grid style=dashed,
            xtick = data
        ]

        \addplot[color=red, fill=red]
            coordinates {(MSE+SSIM,6.0)(MSE+TV,4.0)(MSE+TV+SSIM, 3.0)};
        
        \addplot[color=cyan, fill=cyan]
            coordinates {(MSE+SSIM,4.0)(MSE+TV,2.3)(MSE+TV+SSIM, 2.1)};
            
        \legend{Translation error (cm),Rotation error (°)}
        \end{axis}
        \end{tikzpicture}
    }
    \caption{\textbf{Impact of additional reconstruction losses on localization accuracy.} Translation and orientation error for several combinations of loss terms.}
    \label{fig:abl_reconstruction}
\end{figure}
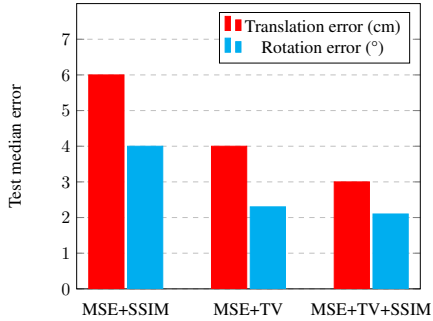

\paragraph{Reconstruction losses.} We evaluated the benefits of the $\mathcal{L}_{SSIM}$ and $\mathcal{L}_{TV}$ terms of the loss function on the localization accuracy on Figure~\ref{fig:abl_reconstruction}. On the Heads scene, the error is 3cm/2.3° with the proposed loss, 4cm/2.1° without $\mathcal{L}_{SSIM}$ and 6cm/4.0° without $\mathcal{L}_{TV}$. These terms actually improve the localization accuracy because they help to recover the correct scene geometry.

\section{Limitations and Future Work} \label{sec_limitations}

\paragraph{Scalability.} Similar to other Neural Scene Representations, our Neural Field struggles to represent large scale maps, such as the one used in autonomous driving, with a single radiance field instance. The current best solution, proposed by Block-NeRF~\cite{tancik2022blocknerf}, is to split the environment into several smaller neural fields and enforce consistency at their boundaries. This solution is successful at a city-scale and could be implemented in our method for large scale localizatoin.

\paragraph{Localization pipeline.} The proposed localization algorithm could be improved in many ways. Dense features matching could be performed by learning-based approaches~\cite{GOCor_Truong_2020, Berton_ICCV_2021, Edstedt_2023_CVPR} instead of simple heuristics. Resulting 2D-3D matchs could be improved by co-visibility filtering~\cite{activesearch, Panek2022ECCV}. Finally, the estimated camera pose could be optimized by direct features alignment, similar to GN-Net~\cite{von2020gn} and PixLoc~\cite{sarlin2021pixloc}. The contribution of this paper lies in the learning of descriptors in a neural renderer, and this proposal can be used as a backbone for different and more advanced localization solutions.

\section{Conclusion} \label{sec_conclusion}

We propose CROSSFIRE; a new way to learn and represent visual localization maps based on neural radiance fields. The proposed formulation has the advantage of densely representing local features of a scene in a compact way, and to be more robust to lightning changes than photometric alignment. We demonstrate that the non-supervised learned local features, which are specialized on the target area, perform better than related supervised techniques that use pre-trained features. The proposed implicit representation can serve as a backbone to more advanced features matching pipelines and should be compatible with future improvements in the neural rendering field that could enable to scale these models to larger scenes and yield better localization accuracy by improving further the quality of the learned scene geometry. We believe that replacing classical data structures by implicit scenes representations is an exciting research direction for the whole area of 3D computer vision as it enables to store dense information in a compact representation.

\section*{Acknowledgments}

We thank Fabien Moutarde, Sascha Hornauer, Quentin Herau and Weichao Qiu for useful research discussions. 

{\small
\bibliographystyle{ieee_fullname}
\bibliography{egbib}
}

\end{document}